\documentclass[10pt,twocolumn,letterpaper]{article}

\usepackage{iccv}
\usepackage{times}
\usepackage{epsfig}
\usepackage{graphicx}
\usepackage{amsmath}
\usepackage{amssymb}

\usepackage{array,multirow,graphicx}
\usepackage{amsmath}
\usepackage{amssymb}
\usepackage{color}
\usepackage{makecell}
\usepackage{hhline}
\usepackage{multicol}
\usepackage{multirow}
\usepackage{arydshln}
\usepackage{soul}
\usepackage{bbding}
\usepackage{textcomp}
\usepackage[shortlabels]{enumitem}

\definecolor{IRO}{rgb}{1,0,0}

\usepackage[breaklinks=true,bookmarks=false]{hyperref}

\iccvfinalcopy 

\def\iccvPaperID{6691} 

\title{3D Scene Graph: A Structure for Unified Semantics, 3D Space, and Camera}

\author{
Iro Armeni$^{1}$ \;\; Zhi-Yang He$^{1}$ \;\; JunYoung Gwak$^{1}$ \;\; Amir R. Zamir$^{1,2}$ \\
Martin Fischer$^{1}$ \;\;  Jitendra Malik$^{2}$ \;\; Silvio Savarese$^{1}$\\ \vspace{-3mm}\\
$^1$ Stanford University \;\; $^2$ University of California, Berkeley\\ \vspace{-1mm}\\
\textcolor{blue}{\url{http://3dscenegraph.stanford.edu}}
}

\begin{document}
\maketitle

\begin{abstract}
A comprehensive semantic understanding of a scene is important for many applications - but in what space should diverse semantic information (e.g., objects, scene categories, material types, texture, etc.) be grounded and what should be its structure? 
Aspiring to have one unified structure that hosts diverse types of semantics, we follow the Scene Graph paradigm in 3D, generating a 3D Scene Graph. Given a 3D mesh and registered panoramic images, we construct a graph that spans the entire building and includes semantics on objects (e.g., class, material, and other attributes), rooms (e.g., scene category, volume, etc.) and cameras (e.g., location, etc.), as well as the relationships among these entities.

However, this process is prohibitively labor heavy if done manually. To alleviate this we devise a semi-automatic framework that employs existing detection methods and enhances them using two main constraints: I. framing of query images sampled on panoramas to maximize the performance of 2D detectors, and II. multi-view consistency enforcement across 2D detections that originate in different camera locations.
\end{abstract}

\begin{figure}
    \centering
    \includegraphics[width=.99\columnwidth]{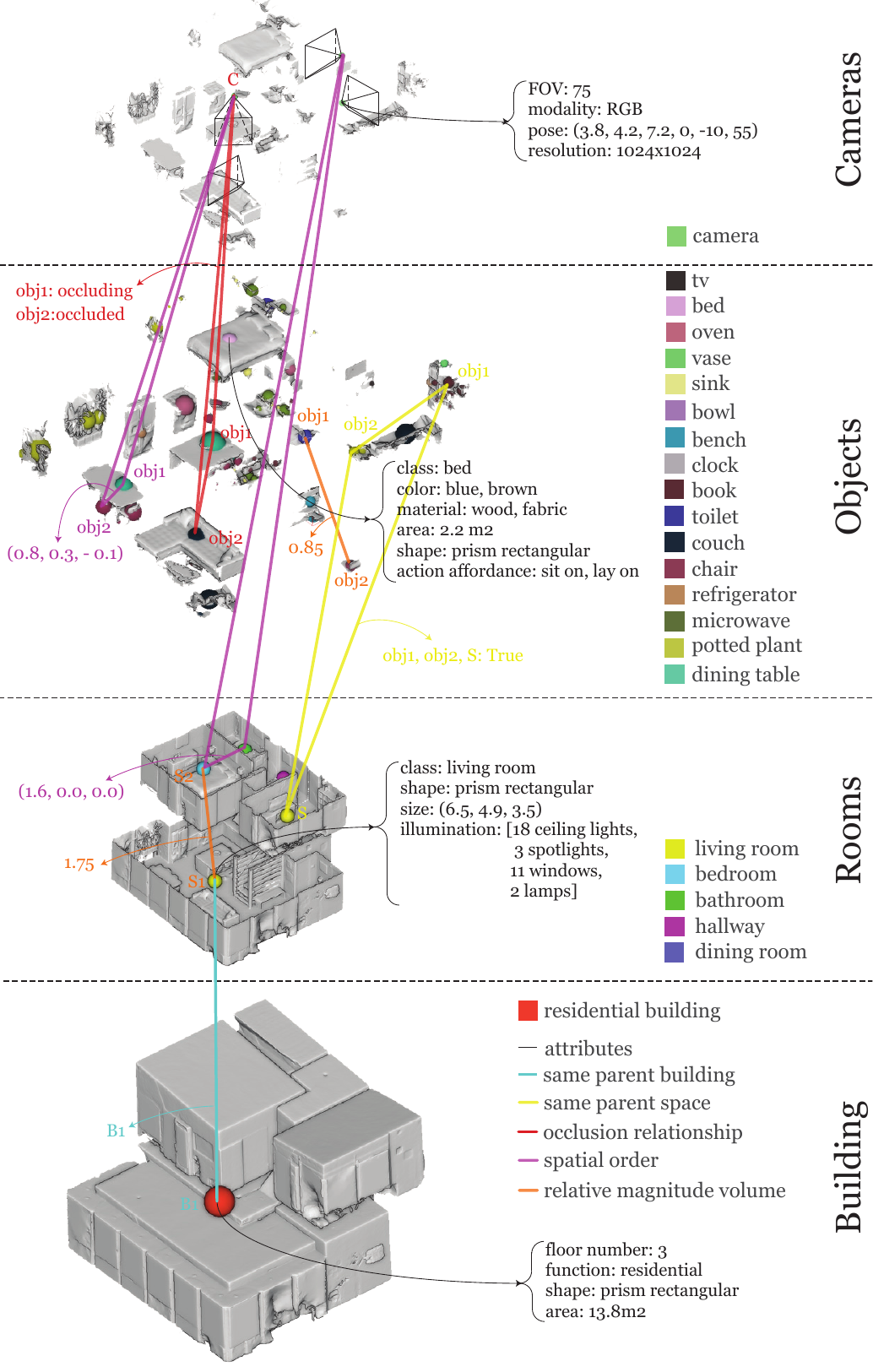}
    \caption{\textbf{3D Scene Graph:} It consists of 4 layers, that represent semantics, 3D space and camera. Elements are nodes in the graph and have certain attributes. Edges are formed between them to denote relationships (e.g., occlusion, relative volume, etc.).}
    \label{fig:3dscenegraph}
    \vspace{-10pt}
\end{figure}

\section{Introduction~\label{sec:intro}}

Where should semantic information be grounded and what structure should it have to be most useful and invariant? This is a fundamental question for a content that preoccupies a number of domains, such as Computer Vision and Robotics. There is a clear number of components in play: geometry of the objects and space, categories of the entities therein, and the viewpoint from which the scene is being observed (i.e. the camera pose).

On the \textit{space} where this information can be grounded, the most commonly employed choice is images. However, the use of images for this purpose is not ideal since it presents a variety of weaknesses, such as pixels being highly variant to any parameter change, the absence of an object's entire geometry, and more. An ideal space for this purpose would be at minimum (a) invariant to as many changes as possible, and (b) easily and deterministically connected to various output ports that different domains and tasks require, such as images or videos. To this end, we articulate that 3D space is more stable and invariant, yet connected to images and other pixel and non-pixel output domains (e.g. depth). Hence, we ground semantic information there, and project it to other desired spaces as needed (e.g., images, etc.). Specifically, this means that the information is grounded on the underlying 3D mesh of a building. This approach presents a number of useful values, such as free 3D, amodal, occlusion, and open space analysis. More importantly, semantics can be projected onto any number of visual observations (images and videos) which provides them with annotations without additional cost.

What should be the \textit{structure}? Semantic repositories use different representations, such as object class and natural language captions. The idea of scene graph has several advantages over other representations that make it an ideal candidate. It has the ability to encompass more information than just object class (e.g., ImageNet~\cite{imagenet}), yet it contains more structure and invariance than natural language captions (e.g., CLEVR~\cite{johnson2017clevr}). We augment the basic scene graph structure, such as the one in Visual Genome~\cite{genome}, with essential 3D information and generate a \textit{3D Scene Graph}.

We view 3D Scene Graph as a layered graph, with each level representing a different entity: building, room, object, and camera. More layers can be added that represent other sources of semantic information. Similar to the 2D scene graph, each entity is augmented with several attributes and gets connected to others to form different types of relationships. To construct the 3D Scene Graph, we combine state-of-the-art algorithms in a mainly automatic approach to semantic recognition. Beginning from 2D, we gradually aggregate information in 3D using two constraints: \textit{framing} and \textit{multi-view consistency}. Each constraint provides more robust final results and consistent semantic output.

The contributions of this paper can be summarized as:
\vspace{-5pt}
\begin{itemize}
    \itemsep0.01em
    \item We extend the scene graph idea in~\cite{genome} to 3D space and ground semantic information there. This gives free computation for various attributes and relationships.
    \item We propose a two-step robustification approach to optimizing semantic recognition using imperfect existing detectors, which allows the automation of a mainly manual task.
    \item We augment the Gibson Environment's~\cite{gibson} database with 3D Scene Graph as an additional modality and make it publicly available at \textcolor{blue}{\href{http://3dscenegraph.stanford.edu}{3dscenegraph.stanford.edu}}.
\end{itemize}

\section{Related Work~\label{sec:relatedwork}}

\paragraph{Scene Graph} A diverse and structured repository is Visual Genome~\cite{genome}, which consists of 2D images in the wild of objects and people. Semantic information per image is encoded in the form of a scene graph. In addition to object class and location, it offers attributes and relationships. The nodes and edges in the graph stem from natural language captions that are manually defined. To address naming inconsistencies due to the free form of annotations, entries are canonicalized before getting converted into the final scene graph. In our work, semantic information is generated in an automated fashion - hence significantly more efficient, already standardized, and to a great extent free from human subjectivity. Although using predefined categories can be restrictive, it is compatible with current learning systems. In addition, 3D Scene Graph allows to compute from 3D an unlimited number of spatially consistent 2D scene graphs and provides numerically accurate quantification to relationships. However, our current setup is limited to indoor static scenes, hence not including outdoor related attributes or action-related relationships, like Visual Genome. 

\vspace{-12pt}
\paragraph{Using Scene Graphs}
Following Visual Genome, several works emerged that employ or generate scene graphs. Examples are on scene graph generation~\cite{li2017scene, xu2017scene}, image captioning/description~\cite{img_caption_a,img_caption_b,img_caption_c}, image retrieval~\cite{retrieval_1} and visual question-answering~\cite{qa_1, qa_2}. Apart from vision-language tasks, there is also a focus on relationship and action detection~\cite{relationship, liang2017deep, zhang2017visual}. A 3D Scene Graph will similarly enable, in addition to common 3D vision tasks, others to emerge in the combination of 3D space, 2D-2.5D images, video streams, and language.

\begin{figure*}
    \centering
    \includegraphics[width=.99\textwidth]{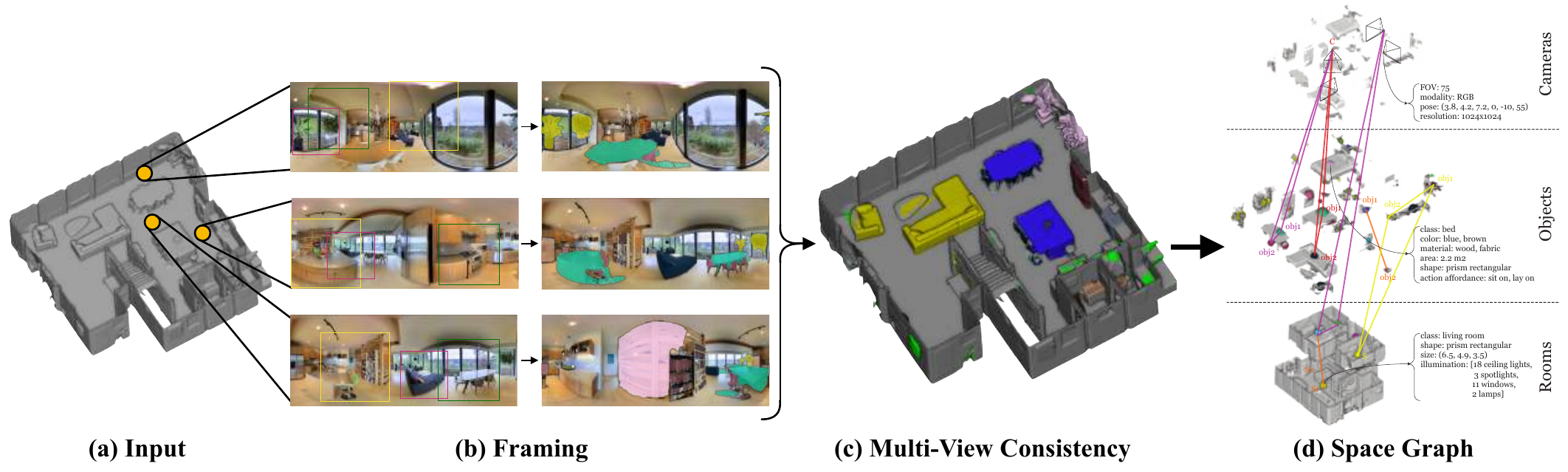}
    \caption{\textbf{Constructing the 3D Scene Graph.} (a) Input to the method is a 3D mesh model with registered panoramic images. (b) Each panorama is densely sampled for rectilinear images. Mask R-CNN detections on them are aggregated back on the panoramas with a weighted majority voting scheme. (c) Single panorama projections are then aggregated on the 3D mesh. (d) These detections become the nodes of 3D Scene Graph. A subsequent automated step calculates the remaining attributes and relationships.}
    \label{fig:method_fig}
    \vspace{-7pt}
\end{figure*}

\vspace{-12pt}
\paragraph{Utilizing Structure in Prediction}
Adding structure to prediction, usually in the form of a graph, has proven beneficial for several tasks. One common application is that of Conditional Random Fields (CRF)~\cite{crf} for \textit{semantic segmentation}, often used to provide globally smooth and consistent results to local predictions~\cite{segcloud, efficient_crf}. In the case of \textit{robot navigation}, employing semantic graphs to abstract the physical map allows the agent to learn by understanding the relationship between semantic nodes independent of the metric space, which results to easier generalization across spaces~\cite{robnav_a}. Graph structures are also commonly used in \textit{human-object interaction} tasks~\cite{hum_obj_a} and other spatio-temporal problems~\cite{structuralRNN}, creating connections among nodes within and across consecutive video frames, hence extending structure to include, in addition to space, also time. Grammars that combine geometry, affordance and appearance have been used toward \textit{holistic scene parsing} in images, where information about the scene and objects is captured in a hierarchical tree structure~\cite{choi2013understanding, zhao2013scene, jiang2018configurable,huang2018holistic}. Nodes represent scene or object components and attributes, whereas edges can represent decomposition (e.g., a scene into objects, etc.) or relationship (e.g., supporting, etc.). Similar to such works, our structure combines different semantic information. However, it can capture global 3D relationships on the building scale and provides greater freedom in the definition of the graph by placing elements in different layers. This removes the need for direct dependencies across them (e.g., between a scene type and object attributes). Another interesting example is that of Visual Memex~\cite{memex} that leverages a graph structure to encode contextual and visual similarities between objects without the notion of categories, with the goal to predict the object class laying under a masked region. Zhu et al.~\cite{affordance_knowledgegraph} used a knowledge base representation for the task of object affordance reasoning that places edges between different nodes of objects, attributes, and affordances. These examples incorporate different types of semantic information in a unified structure for multi-modal reasoning. The above echoes the value of having richly structured information.

\vspace{-13pt}
\paragraph{Semantic Databases}
Existing semantic repositories are fragmented to specific types of visual information, with their majority focusing on object class labels and spatial span/positional information (e.g., segmentation masks/bounding boxes). These can be further sub-grouped based on the visual modality (e.g., RGB, RGBD, point clouds, 3D mesh/CAD models, etc.) and content scene (e.g., indoor/outdoor, object only, etc.). Among them, a handful provides multimodal data grounded on 3D meshes (e.g., 2D-3D-S~\cite{2d3ds}, Matterport3D~\cite{mp3d}). The Gibson database~\cite{gibson}, consists of several hundreds of 3D mesh models with registered panoramic images. It is approximately 35 and 4.5 times larger in floorplan than the 2D-3D-S and Matterport3D datasets respectively, however, it currently lacks semantic annotations. Other repositories specialize on different types of semantic information, such as materials (e.g., Materials in Context Database (MINC)~\cite{MINC}), visual/tactile textures (e.g., Describable Textures Dataset (DTD)~\cite{DTD}) and scene categories (e.g., MIT Places~\cite{MITplaces}).

\vspace{-12pt}
\paragraph{Automatic and Semi-automatic Semantic Detection}
Semantic detection is a highly active field (a detailed overview is out of scope for this paper). The main point to stress is that, similar to the repositories, works are focused on a limited semantic information scope. \textit{Object Semantics} range from class recognition to spatial span definition (bounding box/segmentation mask). One of the most recent works is Mask R-CNN~\cite{maskrcnn}, which provides object instance segmentation masks in RGB images. Other ones with similar output are Blitz-Net~\cite{blitznet} (RGB) and Frustum PointNet~\cite{frustum} (RGB-D).

In addition to detection methods, crowd-sourcing data annotation is a common strategy, especially when building a new repository. Although most approaches focus solely on manual labor, some employ automation to minimize the amount of human interaction with the data and provide faster turnaround. Similar to our approach, Andriluka et al.~\cite{andriluka2018fluid} employ Mask R-CNN trained on the COCO-Stuff dataset to acquire initial object instance segmentation masks that are subsequently verified and updated by users. Polygon-RNN~\cite{castrejon2017annotating, acuna2018efficient} is another machine-assisted annotation tool which provides contours of objects in images given user-defined bounding boxes. Both remain in the 2D world and focus on object category and segmentation mask.  

\begin{figure*}
    \centering
    \includegraphics[width=\textwidth]{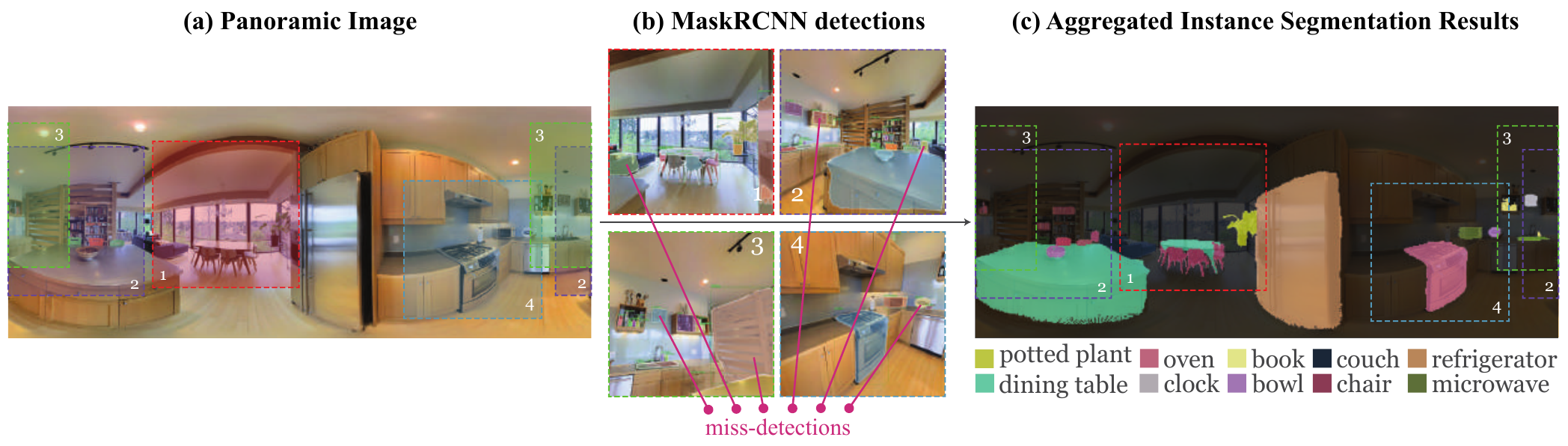}
    \caption{\textbf{Framing:} Examples of sampled rectilinear images using the framing robustification mechanism are shown in the dashed colored boxes. Detections (b) on individual frames are not error-free (miss-detections are shown with arrows). The errors are pruned out with weighted majority voting to get the final panorama labels.}
    \label{fig:framing}
\end{figure*}

Others employ lower-level automation to accelerate annotations in 3D. ScanNet~\cite{scannet} proposes a web-interface for manual annotation of 3D mesh models of indoor spaces. It begins with an over-segmentation of the scene using a graph-cut based approach. Users are then prompted to label these segments with the goal of object instance segmentation. \cite{sceneNN} has a similar starting point; the resulting over-segments are further grouped into larger regions based on geometry and appearance cues. These regions are edited by users to get object semantic annotations.  \cite{Labelme3D} employs object segmentation masks and labels from 2D annotations to automatically recover the 3D scene geometry.

Despite the incorporation of automation, the above rely largely on human interaction to achieve sufficiently accurate results.

\section{3D Scene Graph Structure~\label{sec:3dscenegraph}}

The \textbf{input} to our method is the typical output of 3D scanners and consists of 3D mesh models, registered RGB panoramas and the corresponding camera parameters, such as the data in Matterport3D~\cite{mp3d} or Gibson~\cite{gibson} databases. 

The \textbf{output} is the 3D Scene Graph of the scanned space, which we formulate as a four-layered graph (see Figure~\ref{fig:3dscenegraph}). Each layer has a set of \textit{nodes}, each node has a set of \textit{attributes}, and there are \textit{edges} between nodes which represent their relationships. The first layer is the entire \textit{building} and includes the root node for a given mesh model in the graph (e.g., a residential building). The \textit{rooms} of the building compose the second layer of 3D Scene Graph, and each room is represented with a unique node (e.g., a living room). \textit{Objects} within the rooms form the third layer (e.g., a chair or a wall). The final layer introduces \textit{cameras} as part of the graph: each camera location is a node in 3D and a possible observation (e.g., an RGB image) is associated with it.\\\textit{Attributes}: Each building, room, object and camera node in the graph - from now on referred to as element - has a set of attributes. Some examples are the object class, material type, pose information, and more.\\\textit{Relationships}: Connections between elements are established with edges and can span within or across different layers (e.g., object-object, camera-object-room, etc.). A full list of the attributes and relationships is in Table~\ref{tab:3dscenegraph_params}.

\section{Constructing the 3D Scene Graph~\label{sec:3dscenegraph_construct}}

To construct the 3D Scene Graph we need to identify its elements, their attributes, and relationships. Given the number of elements and the scale, annotating the input RGB and 3D mesh data with object labels and their segmentation masks is the major labor bottleneck of constructing the 3D Scene Graph. Hence the primary focus of this paper is on addressing this issue by presenting an automatic method that uses existing semantic detectors to bootrstap the annotation pipeline and minimize human labor. An overview of the pipeline is shown in Figure~\ref{fig:method_fig}. In our experiments (Section~\ref{sec:experiments}), we used the best reported performing Mask R-CNN network~\cite{maskrcnn} and got results only for detections with a confidence score of 0.7 or higher. However, since detection results are imperfect, we propose two robustification mechanisms to increase their performance, namely \textit{framing} and \textit{multi-view consistency}, that operate on the 2D and 3D domains respectively. 

\begin{table}[ht]
    \centering
    \caption{\textbf{3D Scene Graph Attributes and Relationships.} For a detailed description see \textcolor{blue}{\href{http://3dscenegraph.stanford.edu/images/supp_mat.pdf}{supplementary material}}~\cite{suppmat}.}
    \vspace{3mm}
    \resizebox{\columnwidth}{!}{
      \begin{tabular}{p{1.5cm}p{6cm}p{6cm}}
          \makecell{\textbf{Elements}} & \makecell{\textbf{Attributes}} & \makecell{\textbf{Relationships}} \\ 
          \parbox[t]{1mm}{\rotatebox[origin=c]{0}{\textit{\textcolor[rgb]{1,0,0}{Object (O)}}}}	& \makecell{Action Affordance, Class, Floor Area,\\
                                    ID, Location, Material, \\
                                    Mesh Segmentation, Size,\\
                                    Texture, Volume, Voxel Occupancy} & \makecell{Amodal Mask \textit{(O,C)}, Parent Space \textit{(O,R)}, \\ 
                                    Occlusion Relationship \textit{(O,O,C)}, \\
                                    Same Parent Room \textit{(O,O,R)}, Spatial Order \textit{(O,O,C)} \\ 
                                    Relative Magnitude \textit{(O,O)}} \\ \hdashline 
          \parbox[t]{1mm}{\rotatebox[origin=c]{0}{\textcolor[rgb]{0,0,1}{\textit{Room (R)}}}}	& \makecell{Floor Area, ID, Location, \\
                            Mesh Segmentation, Scene Category, \\
                            Size, Volume, Voxel Occupancy} & \makecell{Spatial Order \textit{(R,R,C)}, Parent Building \textit{(R,B)}, \\
                            Relative Magnitude \textit{(R,R)}} \\ \hdashline
          \parbox[t]{1mm}{\rotatebox[origin=c]{0}{\textcolor[rgb]{0,1,0}{\textit{Building (B)}}}}	& \makecell{Area, Building Reference Center, \\
                                  Function, ID, Number of Floors, \\
                                  Size, Volume} &  \\ \hdashline
          \parbox[t]{1mm}{\rotatebox[origin=c]{0}{\textcolor[rgb]{1,0,1}{\textit{Camera(C)}}}}	& \makecell{Field Of View, ID, Modality, \\
                                  Pose, Resolution} & \makecell{Parent Space {(C,R)}} \\
         \makecell{\multirow{3}{13.5cm}{\textbf{Note:} For Relationships, \textit{(X,Y)} means that it is between elements \textit{X} and \textit{Y}. It can also be among a triplet of elements \textit{(X,Y,Z)}.  Elements can belong to the same category (e.g., O,O - two Objects) or different ones (e.g., O,C - an Object and a Camera).}}
      \end{tabular}
    }
    \label{tab:3dscenegraph_params}
\end{table}

\vspace{-10pt}
\paragraph{Framing on Panoramic Images}
2D semantic algorithms operate on rectilinear images and one of the most common errors associated with their output is incorrect detections for partially captured objects at the boundaries of the images. When the same objects are observed from a slightly different viewpoint that places them closer to the center of the image and does not partially capture them, the detection accuracy is improved. Having RGB panoramas as input gives the opportunity to formulate a framing approach that samples rectilinear images from them with the objective to maximize detection accuracy. This approach is summarized in Figure~\ref{fig:framing}. It utilizes two heuristics: (a) placing the object at the center of the image and (b) having the image properly zoomed-in around it to provide enough context. We begin by densely sampling rectilinear images on the panorama with different yaw ($\psi$), pitch ($\theta$) and Field of View (FoV) camera parameters, with the goal of having at least one image that satisfies the heuristics for each object in the scene:
$$\psi = [-180^{\circ}, 180^{\circ}, 15^{\circ}], \theta = [-15^{\circ}, 15^{\circ}, 15^{\circ}]$$\vspace{-10pt}
$$FoV = [75^{\circ}, 105^{\circ}, 15^{\circ}]$$
This results in a total of 225 images of size 800 by 800 pixels per panorama and serves as the input to Mask-RCNN. To prune out imperfections in the rectilinear detection results, we aggregate them on the panorama using a weighted voting scheme where the weights take into account: the predictions' confidence score and the distance of the detection from the center of the image. In specific, we compute weights per pixel for each class as follows:
\begin{equation*}
w_{i,\lambda} = \sum_{j, L_{d_{ij}}=\lambda} \frac{S_{d_{ij}}}{\lVert C_{d_{ij}}-C_{j} \rVert}
\end{equation*}

\noindent where $w_{i,\lambda}$ is the weight of panorama pixel $i$ for class $\lambda$, $L_{d_{ij}}$ is the class of detection $d_{ij}$ for $i$ in rectilinear frame $j$, $S_{d_{ij}}$ is the confidence score  and $C_{d_{ij}}$ is the center pixel location for the detection, and $C_{j}$ is the center of $j$. Given these weights, we compute the highest scoring class per pixel. However, performing the aggregation on individual pixels can result to local inconsistencies, since it disregards information on which pixels could belong to an object instance. Thus, we look at each rectilinear detection and use the highest scoring classes of the contained panorama pixels as a pool of candidates for their final label. We assign the one that is the most prevalent among them. At this stage, the panorama is segmented per class, but not per instance. To address this, we find the per-class connected components; this gives us instance segmentation masks.

\begin{figure}
    \centering
    \includegraphics[width=\columnwidth]{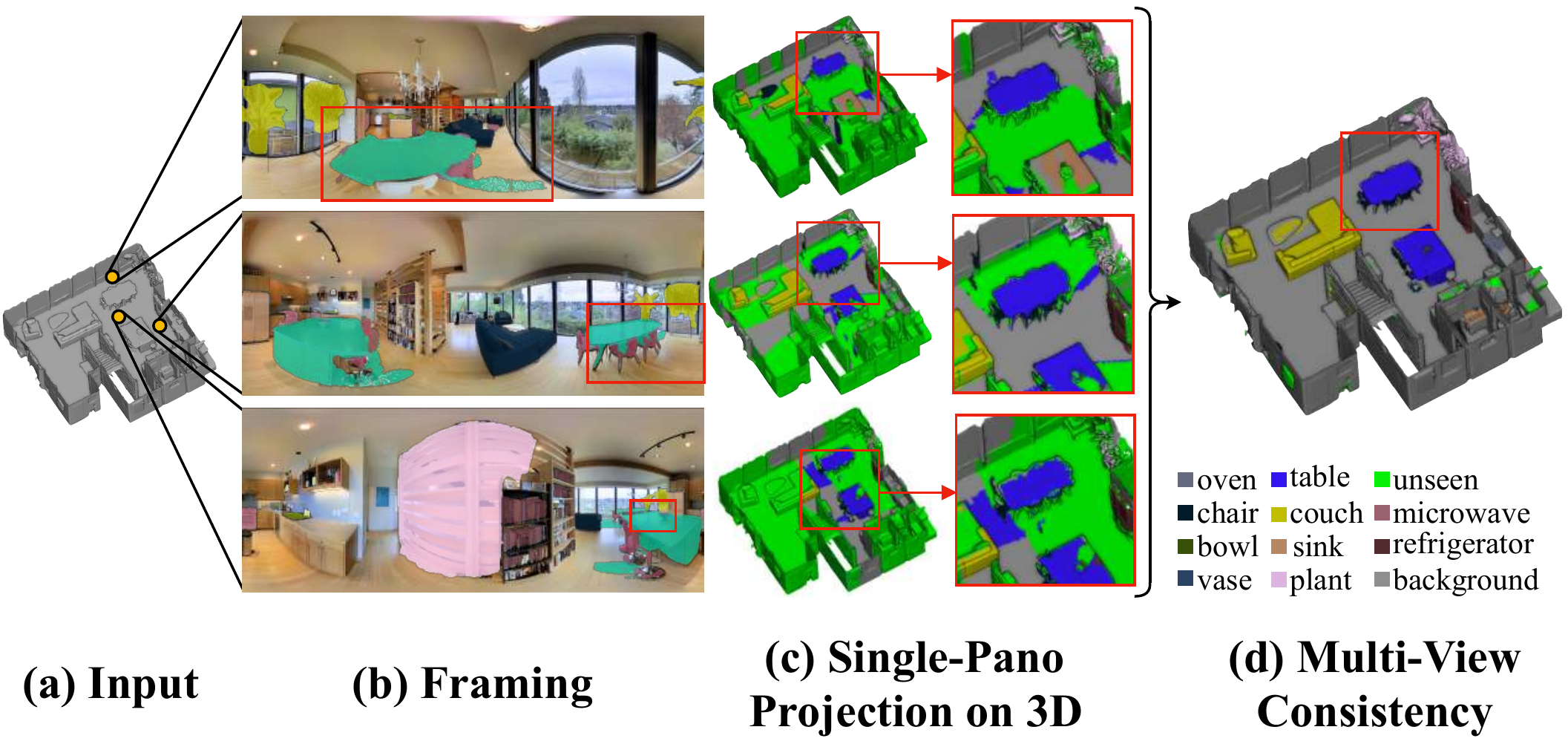}
    \vspace{1mm}
    \caption{\textbf{Multi-view consistency:} Semantic labels from different panoramas are combined on the final mesh via multi-view consistency. Even though the individual projections carry errors from the panorama labels and poor 3D reconstruction/camera registration, observing the object from different viewpoints can fix them.}\label{fig:mvc} 
\end{figure}

\paragraph{Multi-view consistency}~\label{par:meshextract}
With the RGB panoramas registered on the 3D mesh, we can annotate it by projecting the 2D pixel labels on the 3D surfaces. However, a mere projection of a single panorama does not yield accurate segmentation, because of imperfect panorama results (Figure~\ref{fig:mvc}(b)), as well as common poor reconstruction of certain objects or misalignment between image pixels and mesh surfaces (camera registration errors). This leads to labels "leaking" on neighboring objects (Figure~\ref{fig:mvc}(c)). However, the objects in the scene are visible from multiple panoramas, which enables using multi-view consistency to fix such issues. This makes our second robustification mechanism. We begin by projecting all panorama labels on the 3D mesh surfaces. To aggregate the casted votes, we formulate a weighted majority voting scheme based on how close an observation point is to a surface, following the heuristic that the closer the camera to the object, the larger and better visible it is. In specific, we define weights as:

\vspace{-3pt}
\begin{equation*}
w_{i,j} = \frac{\sum_{i,j} \lVert P_{i}-F_{c_{j}} \rVert}{\lVert P_{i}-F_{c_{j}} \rVert}
\end{equation*}

\noindent where $w_{i,j}$ is the weight of a face $F_{j}$ with respect to a camera location $P_{i}$ and $F_{c_{j}}$ is the 3D coordinates of $F_{j}$'s center.

Similar to the framing mechanism, voting is performed on the detection level. We look for label consistency across the group of faces $F_{obj}$ that receives votes from the same object instance in a panorama. We first do weighted majority voting on individual faces to determine the pool of label candidates for $F_{obj}$ as it results from casting all panoramas, and then use the one that is most present to assign it to the group. A last step of finding connected components in 3D gives us the final instance segmentation masks. This information can be projected back on the panoramas, hence providing consistent 2D and 3D labels.

 \begin{figure*}[ht]
    \centering
    \begin{tabular}{ccc}
        \includegraphics[width=0.32\textwidth]{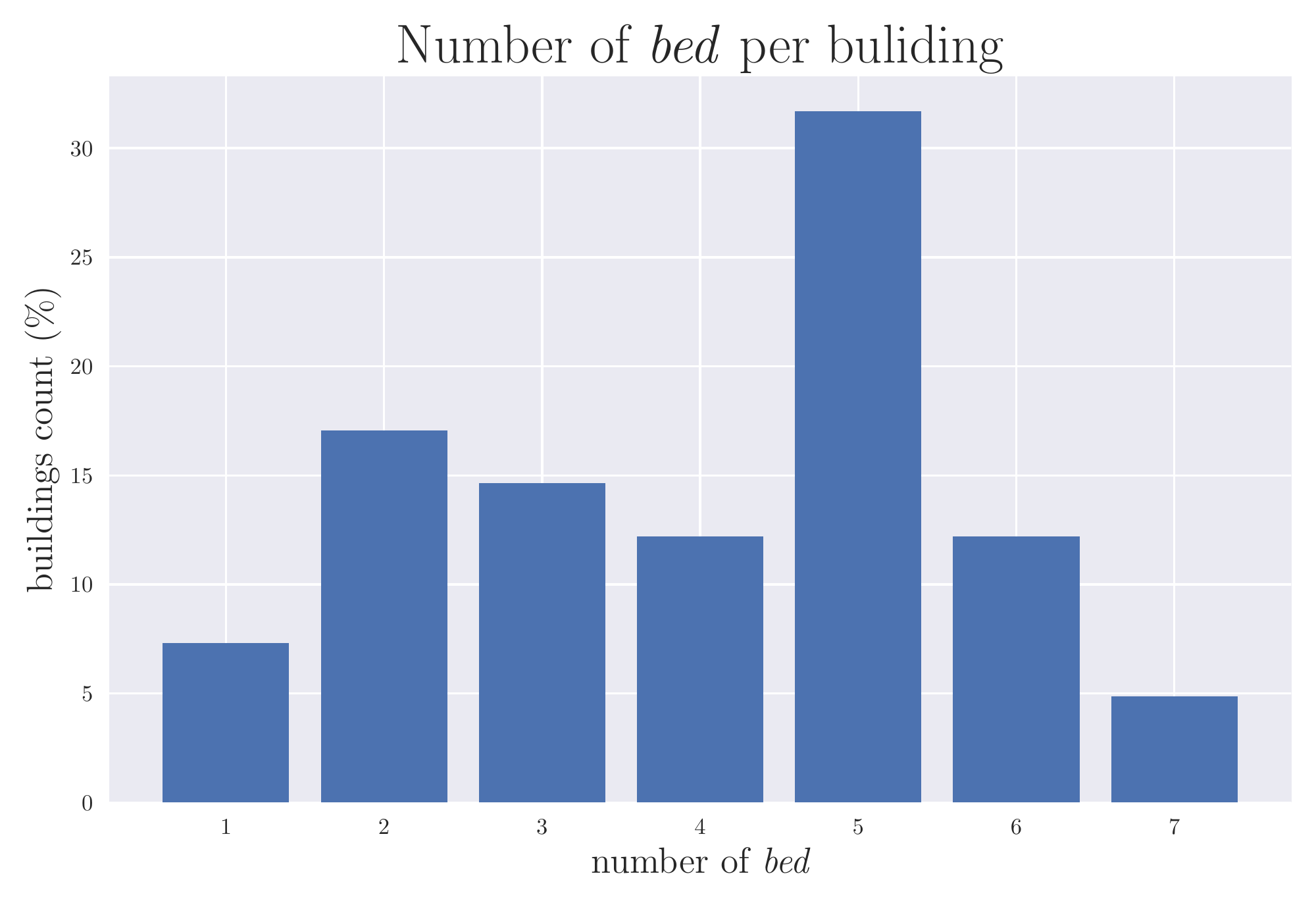}&
        \includegraphics[width=0.32\textwidth]{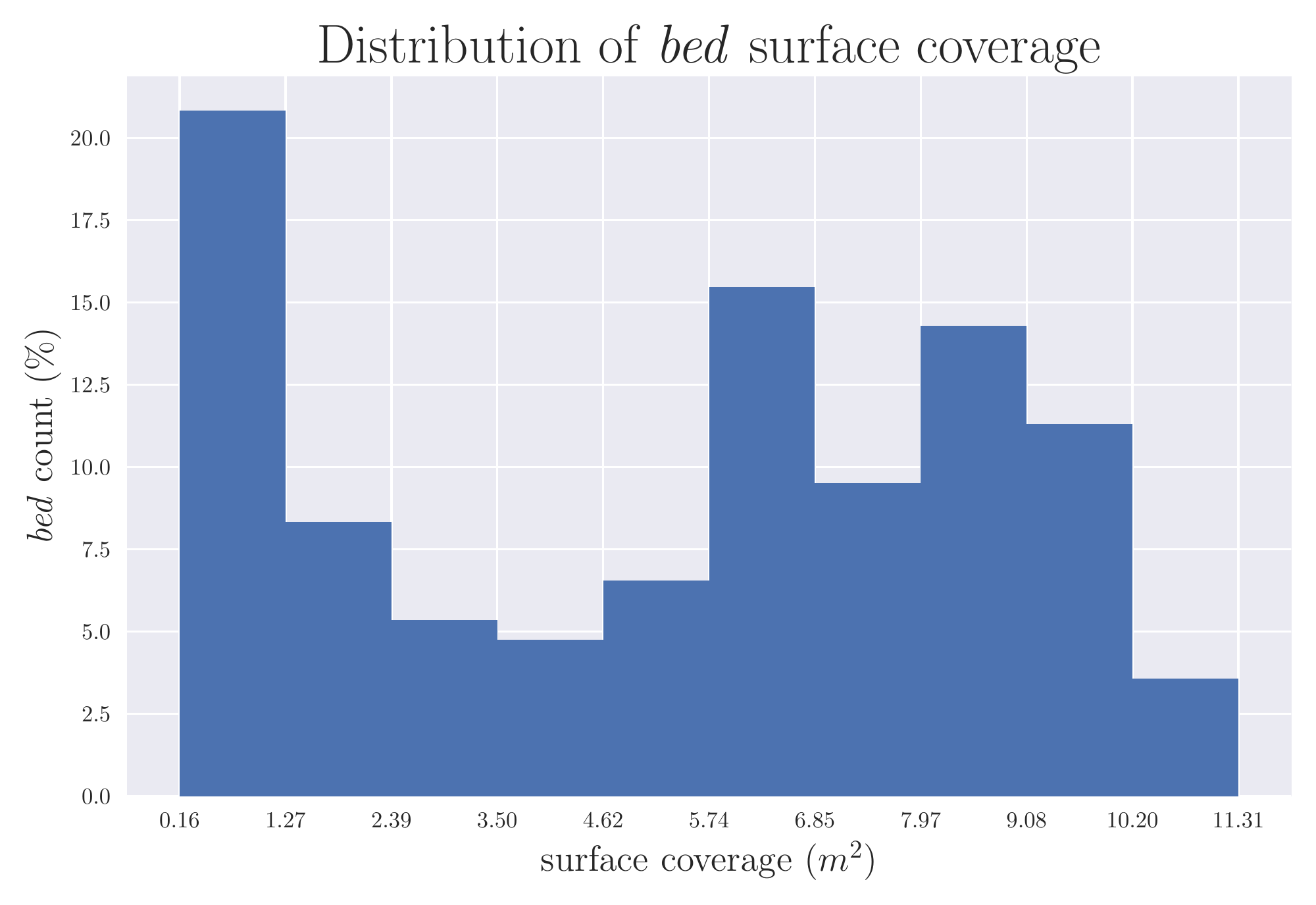}&
        \includegraphics[width=0.32\textwidth]{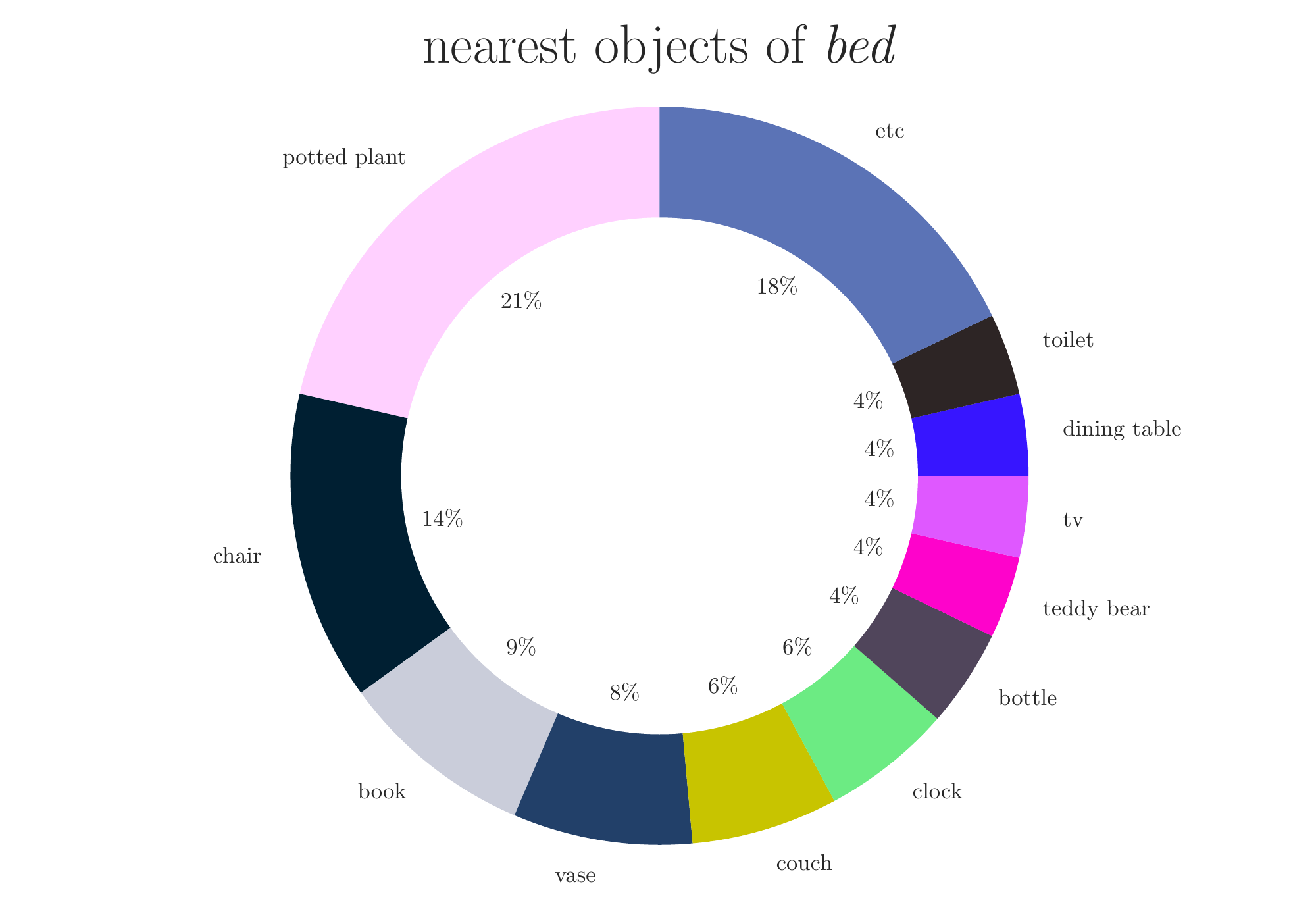}
    \end{tabular}
    \caption{\textbf{Semantic statistics for \textit{bed}:} (a) Number of object instances in buildings. (b) Distribution of its surface coverage. (c) Nearest object instance in 3D space. \textit{(from left to right)}}
    \label{fig:stats_teaser}
\end{figure*}

\subsection{User-in-the-loop verification}
As a final step, we perform manual verification of the automatically extracted results. We develop web interfaces with which users verify and correct them when necessary. Screenshots and more details on this step are offered in the \textcolor{blue}{\href{http://3dscenegraph.stanford.edu/images/supp_mat.pdf}{supplementary material}}~\cite{suppmat}. We crowd-sourced the verification in Amazon Mechanical Turk (AMT). However, we do not view this as a crucial step of the pipeline as the automated results without any verification are sufficiently robust to be of certain practical uses (see Section~\ref{sec:scenegraphprediction} and the \textcolor{blue}{\href{http://3dscenegraph.stanford.edu/images/supp_mat.pdf}{supplementary material}}~\cite{suppmat}). The manual verification is performed mostly for evaluation purposes and forming error-free data for certain research use cases. 

The pipeline consists of two main steps (all operations are performed on rectilinear images). \textit{Verification and editing:} After projecting the final 3D mesh labels on panoramas, we render rectilinear images that show each found object in the center and to its fullest extent, including 20\% surrounding context. We ask users to (a) verify the label of the shown object - if wrong, the image is discarded from the rest of the process; (b) verify the object's segmentation mask; if the mask does not fulfill the criteria, users (c) add a new segmentation mask. \textit{Addition of missing objects:} The previous step refines our automatic results, but there may still be missing objects. We project the verified masks back on the panorama and decompose it in 5 overlapping rectilinear images ($72^{\circ}$ of yaw difference per image). This step (a) asks users if any instance of an object category is missing, and if found incomplete, (b) they recursively add masks until all instances of the object category are masked out. 

\subsection{Computation of attributes and relationships}
The described approach gives as output the object elements of the graph. However, a 3D Scene Graph consists of more element types, as well as their attributes and in-between relationships. To compute them, we use off-the-shelf learning and analytical methods. We find \textit{room} elements using the method in~\cite{bldgparser}. The attribute of \textit{volume} is computed using the 3D convex hull of an element. That of \textit{material} is defined in a manual way since existing networks did not provide results with adequate accuracy. All relationships are a result of automatic computation. For example, we compute the 2D \textit{amodal mask} of an object given a camera by performing ray-tracing on the 3D mesh, and the \textit{relative volume} between two objects as the ratio of their 3D convex hull volumes. For a full description of them and for a video with results see the \textcolor{blue}{\href{http://3dscenegraph.stanford.edu/images/supp_mat.pdf}{supplementary material}}~\cite{suppmat}.

\section{Experiments~\label{sec:experiments}}
We evaluate our automatic pipeline on the Gibson Environment's~\cite{gibson} database. 

\subsection{Dataset Statistics~\label{sec:statistics}}
The Gibson Environment's database consists of 572 full buildings. It is collected from real indoor spaces and provides for each building the corresponding 3D mesh model, RGB panoramas and camera pose information\footnote{For more details visit \textcolor{blue}{\href{http://gibsonenv.stanford.edu/database/}{gibsonenv.stanford.edu/database}}}. We annotate with our automatic pipeline all 2D and 3D modalities, and manually verify this output on Gisbon's tiny split. The semantic categories used come from the COCO dataset~\cite{coco} for objects, MINC~\cite{MINC} for materials, and DTD~\cite{DTD} for textures. A more detailed analysis of the dataset and insights per attributes and relationships is in the \textcolor{blue}{\href{http://3dscenegraph.stanford.edu/images/supp_mat.pdf}{supplementary material}}~\cite{suppmat}. Here we offer an example of semantic statistics for the object class of \textit{bed} (Figure~\ref{fig:stats_teaser}).

\subsection{Evaluation of Automated Pipeline~\label{sec:accuracy}}
We evaluate our automated pipeline both on 2D panoramas and 3D mesh models. We follow the COCO evaluation protocol~\cite{coco} and report the average precision (AP) and recall (AR) for both modalities. We use the best off-the-shelf Mask R-CNN model trained on the COCO dataset. Specifically, we choose Mask R-CNN with Bells \& Whistles from Detectron~\cite{detectron}. According to the model notes, it uses a ResNeXt-152 (32x8d)~\cite{xie2017aggregated} in combination with a Feature Pyramid Network (FPN)~\cite{lin2017feature}. It is pre-trained on ImageNet-5K and fine-tuned on COCO. For more details on implementation and training/testing we refer the reader to Mask R-CNN~\cite{maskrcnn} and Detectron~\cite{detectron}.
 
 \begin{table*}[ht]
     \centering
     \caption{\textbf{Evaluation of the automated pipeline on 2D panoramas and 3D mesh.}  We compute Average Precision (AP) and Average Recall (AR) for both modalities based on COCO evaluation~\cite{coco}. Values in parenthesis represent the absolute difference of the AP of each step with respect to the baseline.}
     \resizebox{\textwidth}{!}{
       \begin{tabular}{c||c:c:c|c:c:c}
         \multirow{3}{*}{\textbf{Method}} & \multicolumn{3}{c}{\textbf{2D}}  & \multicolumn{3}{c}{\textbf{3D}}\\
	 & \multirow{2}{*}{Mask R-CNN} & Ours & Ours &  \multirow{2}{*}{Mask R-CNN} & Ours & Ours \\ 
          &  & Mask R-CNN & Mask R-CNN &   & Mask R-CNN & Mask R-CNN \\ 
          & ~\cite{maskrcnn} & w/ Framing &  w/ Framing + MVC & + Pano Projection & w/ Framing + Pano Projection &  w/ Framing + MVC\\ \hline
         \makecell{\textbf{AP}} & 0.079 & 0.160 \small{(+0.081)} & \textbf{0.485 \small{(+0.406)}} &  0.222 & 0.306 \small{(+0.084)} & \textbf{ 0.409 \small{(+0.187)}} \\
         \makecell{\textbf{AP.50}} & 0.166 & 0.316 \small{(+0.150)} & \textbf{0.610 \small{(+0.444)}} & 0.445 & 0.539 \small{(+0.094)} & \textbf{0.665 \small{(+0.220)}} \\
         \makecell{\textbf{AP.75}} & 0.070 & 0.147 \small{(+0.077)} & \textbf{0.495 \small{(+0.425)}} & 0.191 & 0.322 \small{(+0.131)} & \textbf{0.421 \small{(+0.230)}} \\
         \makecell{\textbf{AR}} & 0.151 & 0.256 \small{(+0.105)} & \textbf{0.537 \small{(+0.386)}} & 0.187  & 0.261 \small{(+0.074)} &  \textbf{0.364 \small{(+0.177)}} \\
       \end{tabular}
       }
     \label{tab:automat3D}
 \end{table*}

\textbf{Baselines:} We compare the following approaches in 2D:
\vspace{-15pt}
\begin{itemize}[leftmargin=\parindent,align=left,labelwidth=\parindent,labelsep=0pt]
    \itemsep0.01em
    \item Mask R-CNN~\cite{maskrcnn}: We run Mask R-CNN on 6 rectilinear images sampled on the panorama with no overlap. The detections are projected back on the panorama.
    \item Mask R-CNN with Framing: The panorama results here are obtained from our first robustification mechanism.
    \item Mask R-CNN with Framing and Multi-View Consistency (MVC) - ours: This is our automated method. The panorama results are obtained after applying both robustification mechanisms.
\end{itemize}

\begin{figure}
    \centering
    \includegraphics[width=\columnwidth]{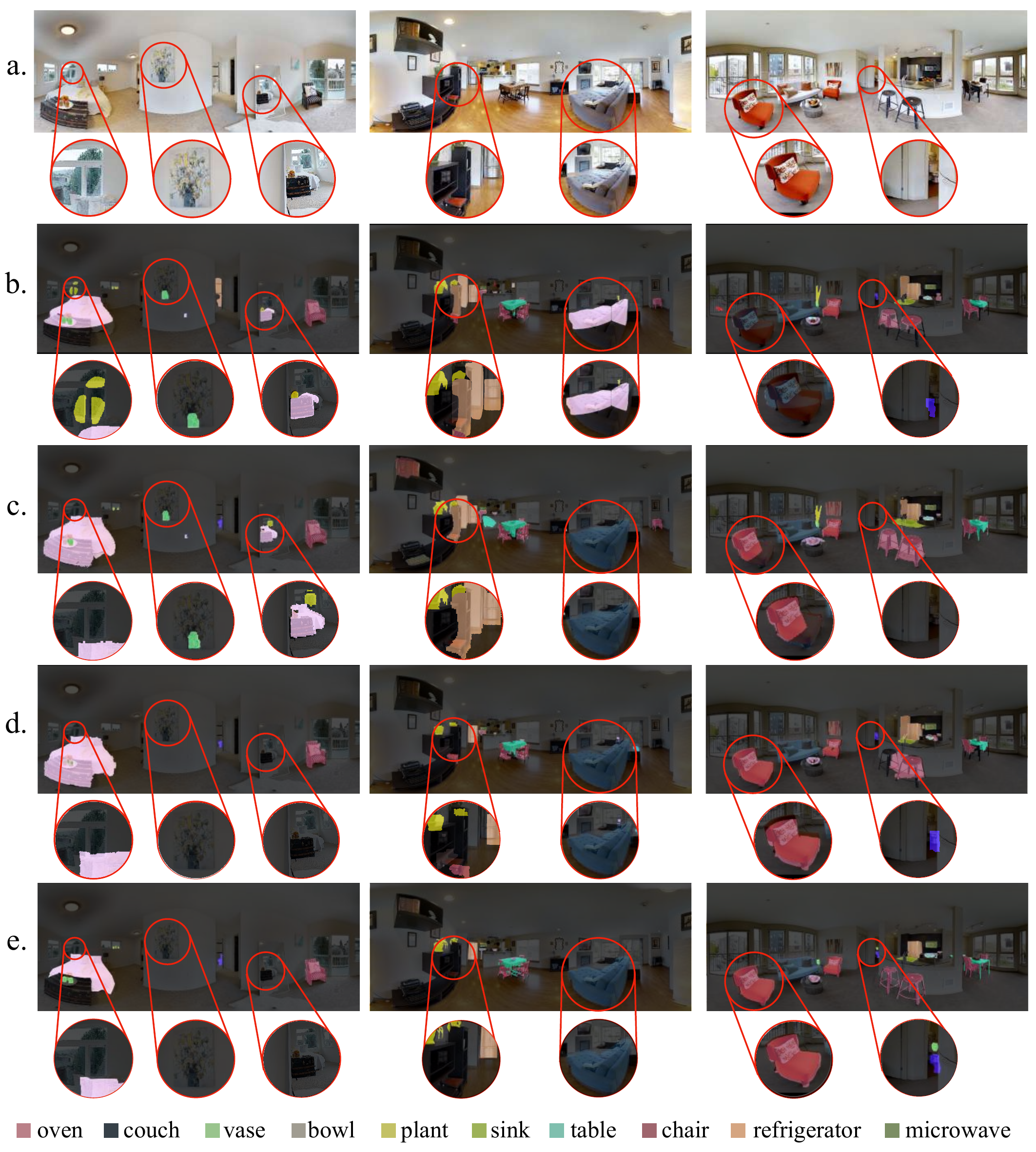}
    \caption{\textbf{Detection results on panoramas:} (a) Image, (b) Mask R-CNN~\cite{maskrcnn}, (c) Mask R-CNN w/ Framing, (d) Mask R-CNN w/ Framing and Multi-View Consistency (our final results), (e) Ground Truth (best viewed on screen). For larger and additional visualizations see the \textcolor{blue}{\href{http://3dscenegraph.stanford.edu/images/supp_mat.pdf}{supplementary material}}~\cite{suppmat}.}
    \label{fig:pano}
\end{figure}

\noindent And these in 3D:
\vspace{-5pt}
\begin{itemize}[leftmargin=\parindent,align=left,labelwidth=\parindent,labelsep=0pt]
    \itemsep0.01em
    \item Mask R-CNN~\cite{maskrcnn} and Pano Projection: The panorama results of Mask R-CNN are projected on the 3D mesh surfaces with simple majority voting per face.
    \item Mask R-CNN with Framing and Pano Projection: The panorama results from our first mechanism follow a similar 2D-to-3D projection and aggregation process.
    \item Mask R-CNN with Framing and Multi-View Consistency (MVC) - ours: This is our automated method.
\end{itemize}

\begin{figure}
    \centering
    \includegraphics[width=\columnwidth]{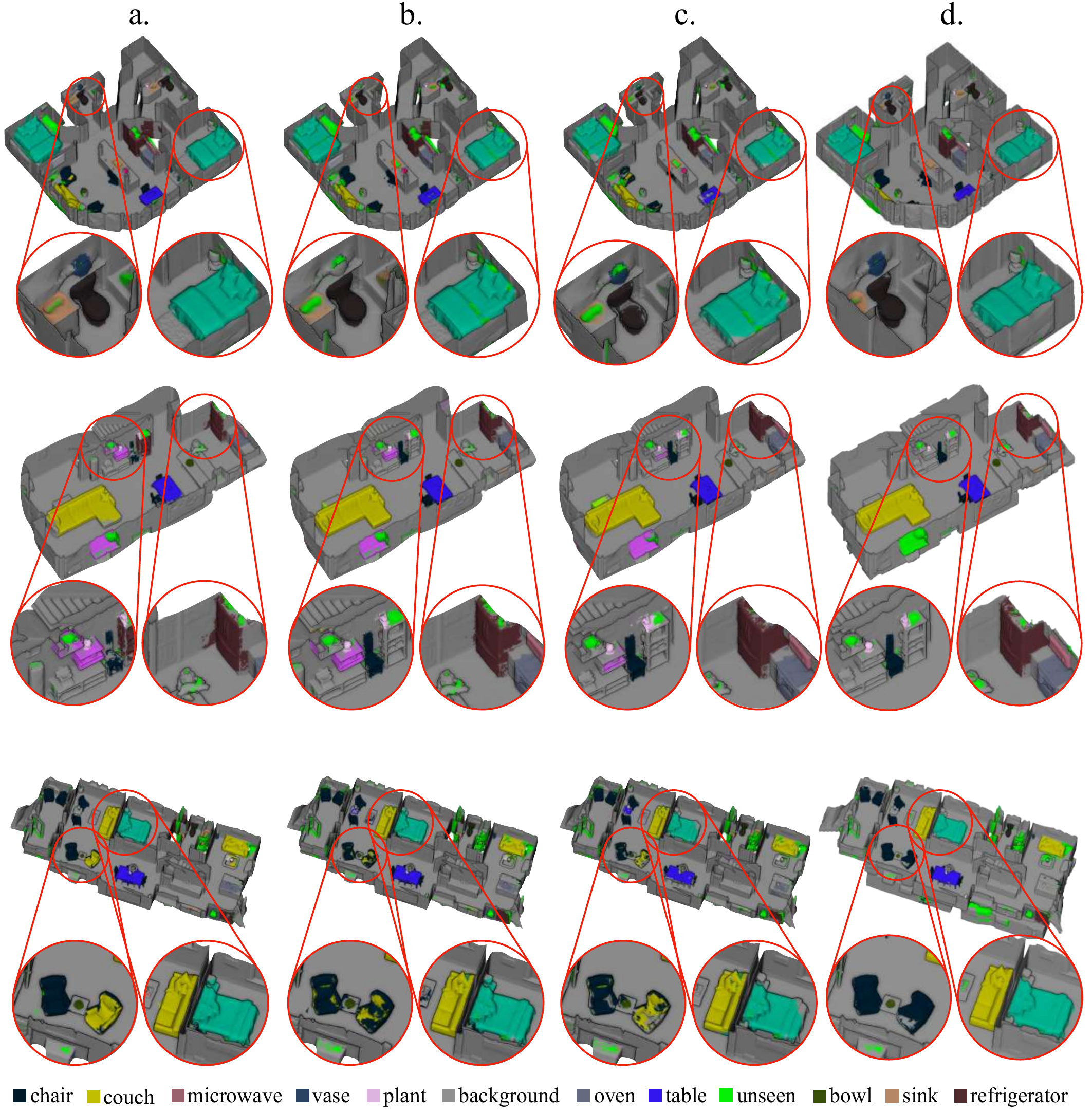}
    \caption{\textbf{3D detection results on mesh:} (a) Mask R-CNN~\cite{maskrcnn} + Pano Projection, (b) Mask R-CNN w/ Framing + Pano Projection, (c) Mask R-CNN w/ Framing and Multi-View Consistency (our final results), (d) Ground Truth (best viewed on screen). For larger and additional visualizations see \textcolor{blue}{\href{http://3dscenegraph.stanford.edu/images/supp_mat.pdf}{supplementary material}}~\cite{suppmat}.}
    \label{fig:mesh}
    \vspace{-4pt}
\end{figure}

As shown in Table~\ref{tab:automat3D}, each mechanism in our approach contributes an additional boost in the final accuracy. This is also visible in the qualitative results, with each step further removing erroneous detections. For example, in the first column of Figure~\ref{fig:pano}, Mask R-CNN (b) detected the trees outside the windows as \textit{potted plants}, a \textit{vase} on a painting and a \textit{bed} reflection in the mirror. Mask R-CNN with framing (c) was able to remove the tree detections and recuperate a missed \textit{toilet} that is highly occluded. Mask R-CNN with framing and multi-view consistency (d) further removed the painted vase and bed reflection, achieving results very close to the ground truth. Similar improvements can be seen in the case of 3D (Figure~\ref{fig:mesh}). Even though they might not appear as large quantitatively, they are crucial for getting consistent 3D results with most changes relating to consistent local regions and better object boundaries. 

\textbf{Human Labor:} We perform a user study to associate detection performance with human labor (hours spent). The results are in Table~\ref{tab:human_labor}. Note that the hours reported for the fully manual 3D annotation~\cite{bldgparser} are computed for 12 object classes (versus 62 in ours) and for an \textit{expert} 3D annotator (versus non-skilled labor in ours).

\vspace{-8pt}
\begin{table}[ht]
    \centering
    \caption{\textbf{Mean time spent by human annotators per model.} Each step is done by 2 users independently for cross checking.}
    \vspace{1mm}
    \resizebox{\columnwidth}{!}{
        \begin{tabular}{c|cc:cc}
            \multirow{2}{*}{\textbf{Method}} & Ours w/o & Ours w/ & Human & ~\cite{bldgparser} \\
             & human (FA) & human (MV) & only (FM 2D) & (FM 3D) \\ \hline
            \textbf{AP} & 0.389 & 0.97 & 1 & 1 \\
            \textbf{Time (h)} & 0 & 03:18:02 & 12:44:10 & 10:18:06 \\ \hline
            \multicolumn{5}{c}{FA: fully automatic | FM: fully manual | MV: manual verification}\\
        \end{tabular}
    }
    \label{tab:human_labor}
\end{table}

\textbf{Using different detectors:}~\label{subsec:impact} Until this point we have been using the best performing Mask R-CNN network with a 41.5 reported AP on COCO~\cite{maskrcnn}. We want to further understand the behavior of the two robustification mechanisms when using a less accurate detector. To this end, we perform another set of experiments using BlitzNet~\cite{blitznet}, a network with faster inference but worse reported performance on the COCO dataset (AP 34.1). We notice that the results for both detectors provide a similar relative increase in AP among the different baselines  (Table~\ref{tab:rel_ap}). This suggests that the robustification mechanisms can provide similar value in increasing the performance of standard detectors and correct errors, regardless of initial predictions.

\begin{table*}[ht]
    \centering
    \caption{\textbf{AP performance of using different detectors.} We compare the performance of two detectors with 7.4 AP difference in the COCO dataset. Values in parenthesis represent the absolute difference of the AP of each step with respect to the baseline.}
    \resizebox{\textwidth}{!}{
      \begin{tabular}{c||c:c:c|c:c:c}
        \multirow{3}{*}{\textbf{Method}} & \multicolumn{3}{c}{\textbf{2D}}  & \multicolumn{3}{c}{\textbf{3D}}\\
          & \multirow{2}{*}{Detector} & Detector & Detector & Detector & Detector & Detector \\
         & & w Framing &  w/ Framing + MVC &+ Pano projection & w Framing + Pano projection &  w/ Framing + MVC \\\hline 
        \makecell{Mask R-CNN~\cite{maskrcnn}} & 0.079 & 0.160 \small{(+0.081)} & \textbf{0.485 \small{(+0.406)}} & 0.222  & 0.306 \small{(+0.084)} & \textbf{0.409 \small{(+0.187)}} \\
        \makecell{BlitzNet~\cite{blitznet}} & 0.095 & 0.198 \small{(+0.103)} & \textbf{0.284 \small{(+0.189)}}  & 0.076 & 0.165 \small{(+0.089)} & \textbf{0.245 \small{(+0.169)}}
      \end{tabular}
    }
    \label{tab:rel_ap}
\end{table*}

\subsection{2D Scene Graph Prediction~\label{sec:scenegraphprediction}}
So far we focused on the automated detection results. These will next go through an automated step to generate the final 3D Scene Graph and \textit{compute attributes and relationships}. Results on this can be seen in the \textcolor{blue}{\href{http://3dscenegraph.stanford.edu/images/supp_mat.pdf}{supplementary material}}~\cite{suppmat}. We use this output for experiments on 2D scene graph prediction. 

There are 3 standard evaluation setups for 2D scene graphs~\cite{lu2016visual}: (a) \textit{Scene Graph Detection:} Input is an image and output is bounding boxes, object categories and predicate (relationship) labels; (b) \textit{Scene Graph Classification:} Input is an image and ground truth bounding boxes, and output is object categories and predicate labels; (c) \textit{Predicate Classification:} Input is an image, ground truth bounding boxes and object categories, and output is the predicate labels. In contrast to Visual Genome where only sparse and instance-specific relationships exist, our graph is dense, hence some of the evaluations (e.g., relationship detection) are not applicable. We focus on \textit{relationship classification} and provide results on: (a) \textit{spatial order} and (b) \textit{relative volume} classification, as well as on (c) \textit{amodal mask segmentation} as an application of the occlusion relationship. 

\vspace{-10pt}
\paragraph{Spatial Order:} Given an RGB rectilinear image and the (visible) segmentation masks of an object pair, we predict if the query object is in front/behind, to the left/right of the other object. We train a ResNet34 using the segmentation masks that were automatically generated by our method, and use the medium Gibson data split. The baseline is Statistically Informed Guess extracted from the training data.

\begin{figure}
    \centering
    \includegraphics[width=.95\columnwidth]{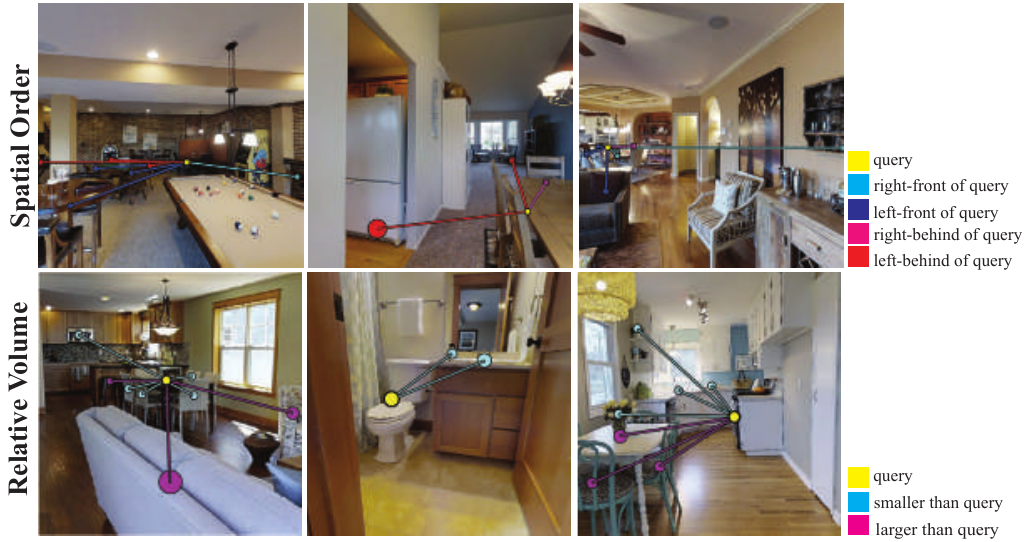}
    \caption{\textbf{Classification results of scene graph relationships with \textit{Ours}.} Each query object is represented with a yellow node. Edges with other elements in the scene showcase a specific relationship. Each relationship type is illustrated with a dedicated color. Color definition is at the right-end of each row. \textbf{Top: \textit{Spatial Order}, Bottom: \textit{Relative Volume}}  (best viewed on screen)}
    \label{fig:rels}
\end{figure}

\noindent \textbf{Relative Volume:} We follow the same setup and predict whether the query object is smaller or larger in volume than the other object. Figure~\ref{fig:rels} shows results of predictions for both tasks, whereas quantitative evaluations are in Table~\ref{tab:mAP}.

\begin{table}[ht]
    \centering
    \caption{\textbf{Mean AP for SG Predicate Classification.}}
    \vspace{1mm}
      \begin{tabular}{c|c|c}
    SG Predicate & Baseline & Ours\\ \hline
        Spatial Order & 0.255 & \textbf{0.712}\\
        Relative Volume  & 0.555 & \textbf{0.820}
      \end{tabular}
     \label{tab:mAP}
\end{table}

\begin{figure}
    \centering
    \includegraphics[width=\columnwidth]{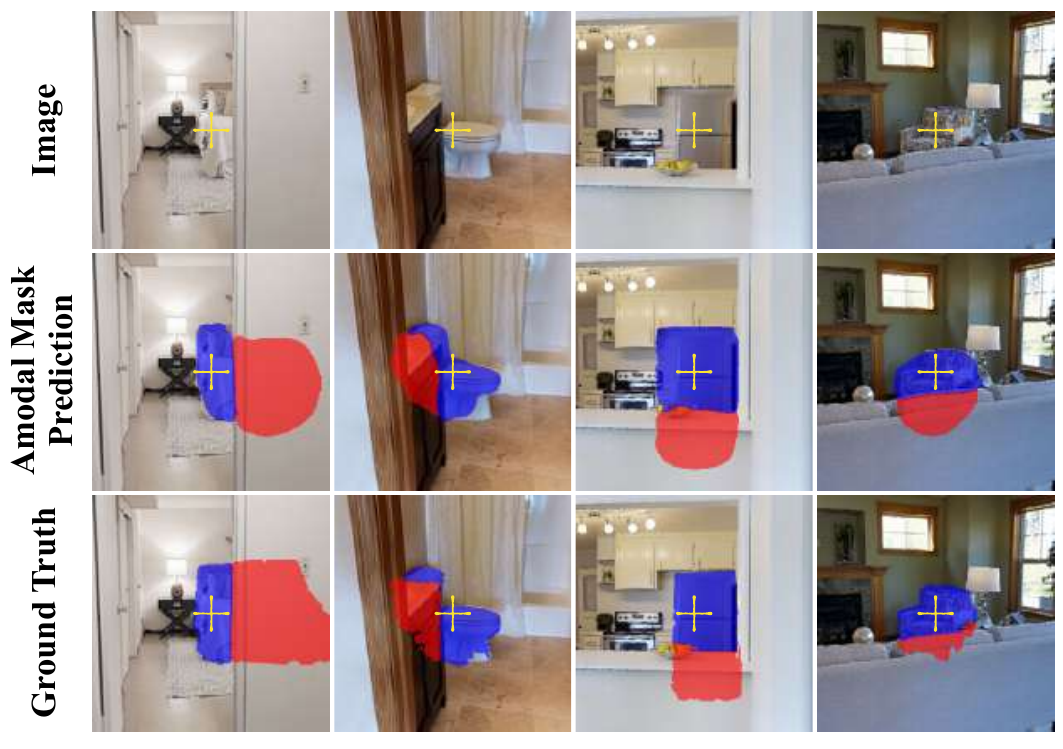}
    \caption{\textbf{Results of amodal mask segmentation with \textit{Ours}.} We predict the visible and occluded parts of the object in the center of an image (we illustrate the center with a cross) \textcolor{blue}{blue: visible}, \textcolor{red}{red: occluded}.}
    \label{fig:amodal}
\end{figure}

\noindent \textbf{Amodal Mask Segmentation:} We predict the 2D amodal segmentation of an object partially occluded by others given a camera location. Since our semantic information resides in 3D space, we can infer the full extents of object occlusions without additional annotations and in a fully automatically way, considering the difficulties of data collection in previous works~\cite{li2016amodal,zhu2017semantic,ehsani2018segan}. We train a U-Net~\cite{ronneberger2015u} agnostic to semantic class, to predict per-pixel segmentation of visible/occluded mask of an object centered on an RGB image (\textit{Amodal Prediction (Ours)}). As baselines, we take an average of amodal masks (a) over the training data (\textit{Avg. Amodal Mask}) and (b) per-semantic class assuming its perfect knowledge at test time (\textit{Avg. Class Specific Amodal Mask}). More information on data generation and experimental setup is in the \textcolor{blue}{\href{http://3dscenegraph.stanford.edu/images/supp_mat.pdf}{supplementary material}}~\cite{suppmat}. We report f1-score and intersection-over-union as a per-pixel classification of three semantic classes (empty, occluded, and visible) along with the macro average (Table~\ref{tab:amodal}). Although the performance gap may not look significant due to a heavy bias of empty class, our approach consistently shows significant performance boost in predicting occluded area, demonstrating that it successfully learned amodal perception unlike baselines (Figure~\ref{fig:amodal}).

\begin{table}[]
    \centering
    \caption{\textbf{Amodal mask segmentation quantitative results.}}
    \vspace{1mm}
    \resizebox{\columnwidth}{!}{
        \begin{tabular}{c|ccc:c}
            f1-score & empty & occluded & visible & avg \\ \hline
            Avg. Amodal Mask & 0.934 & 0.000 & 0.505 & 0.479 \\
            Avg. Class Specific Amodal Mask& 0.939 & 0.097 & 0.599 & 0.545 \\
            Amodal Prediction (Ours) & \textbf{0.946} & \textbf{0.414} & \textbf{0.655} & \textbf{0.672} \\ \hline \hline
            IoU & empty & occluded & visible & avg \\ \hline
            Avg. Amodal Mask & 0.877 & 0.0   & 0.337 & 0.405 \\
            Avg. Class Specific Amodal Mask  & 0.886 & 0.051 & 0.427 & 0.455 \\
            Amodal Prediction (Ours) & \textbf{0.898} & \textbf{0.261} & \textbf{0.488} & \textbf{0.549}
        \end{tabular}
    }
    \label{tab:amodal}
\end{table}

\section{Conclusion}
We discussed the grounding of multi-modal 3D semantic information in a unified structure that establishes relationships between objects, 3D space, and camera. We find that such a setup can provide insights on several existing tasks and allow new ones to emerge in the intersection of semantic information sources. To construct the 3D Scene Graph, we presented a mainly automatic approach that increases the robustness of current learning systems with \textit{framing} and \textit{multi-view consistency}. We demonstrated this on the Gibson dataset, which 3D Scene Graph results are publicly available. We plan to extend the object categories to include more objects commonly present in indoor scenes, since current annotations tend to be sparse in places.

\vspace{-5pt}
\paragraph{Acknowledgements:} \small{We acknowledge the support of Google (GWNHT), ONR MURI (N00014-16-l-2713), ONR MURI (N00014-14-1-0671), and Nvidia (GWMVU).}


\end{document}